\documentclass[a4paper]{style/svproc}
\usepackage[T1]{fontenc}
\usepackage{graphicx}
\usepackage{dsfont}

\usepackage{graphics} 
\usepackage{amsmath} 
\usepackage{amssymb}  
\usepackage{todonotes}

\usepackage{amsfonts}
\usepackage{mathtools}
\usepackage[linesnumbered,algoruled,boxed,vlined, noend]{algorithm2e}
\usepackage{amsthm}
\usepackage{comment}
\usepackage{cite}
\usepackage{wrapfig}

\usepackage{enumitem}
\usepackage{overpic}
\usepackage{xcolor}

\usepackage{ulem}
\DeclareMathAlphabet{\mathcal}{OMS}{cmsy}{m}{n} 

\DeclarePairedDelimiterX{\norm}[1]{\lVert}{\rVert}{#1}

\newcommand{\rom}[1]{\uppercase\expandafter{\romannumeral #1\relax}}

\newcommand{\rect}{\text{$\mathcal{R}$}}

\newcommand{\config}{\text{$\mathcal{X}$}}

\newcommand{\workspace}{\text{$\mathcal{W}$}}
\newcommand{\objects}{\text{$\mathcal{O}$}}
\newcommand{\target}{\text{$\mathcal{T}$}}
\newcommand{\positions}{\text{$\mathcal{P}$}}
\newcommand{\connectedcomp}{\text{$\mathcal{CC}$}}
\newcommand{\pathregion}{\text{$\Pi$}}
\newcommand{\reward}{\text{$\mathrm{rwd}$}}

\newcommand\red[1]{{\color{black}#1}}
\renewcommand\sout[1]{{}}

{\end{list}}

\usepackage{soul} 
\usepackage[a-2b,mathxmp]{pdfx}[2018/12/22]

\begin{document}
\title{Effective and Robust Non-Prehensile Manipulation via Persistent Homology Guided Monte-Carlo Tree Search \vspace{-.2in} }
%
%
\author{Ewerton R. Vieira\inst{1} \and
Kai Gao\inst{2} \and
Daniel Nakhimovich\inst{2}\and\\
Kostas E. Bekris\inst{2}\and
Jingjin Yu\inst{2}\thanks{This work supported by NSF CCF-2309866. EV is partially supported by Air Force Office of Scientific Research under award numbers FA9550-23-1-0011 and FA9550-23-1-0400.}\vspace{-.1in}}
\authorrunning{Vieira et al.}
\titlerunning{
Persistent Homology Guided
Monte-Carlo Tree Search
}
%
\institute{Dept. of Mathematics, Rutgers, NJ, USA
\and Dept. of Computer Science, Rutgers, NJ, USA \vspace{-.2in}}

\maketitle              
\begin{abstract}
Performing object retrieval in real-world workspaces must tackle challenges including \emph{uncertainty} and \emph{clutter}. One option is to apply prehensile operations, which can be time consuming in highly-cluttered scenarios. On the other hand, non-prehensile actions, such as pushing simultaneously multiple objects, can help to quickly clear a cluttered workspace and retrieve a target object. Such actions, however, can also lead to increased uncertainty as it is difficult to estimate the outcome of pushing operations. The proposed framework in this work integrates topological tools and Monte-Carlo Tree Search (MCTS) to achieve effective and robust pushing for object retrieval. It employs persistent homology to automatically identify manageable clusters of blocking objects without the need for manually adjusting hyper-parameters. Then, MCTS uses this information to explore feasible actions to push groups of objects, aiming to minimize the number of operations needed to clear the path to the target. Real-world experiments using a Baxter robot, which involves some noise in actuation, show that the proposed framework achieves a higher success rate in solving retrieval tasks in dense clutter than alternatives. Moreover, it produces solutions with few pushing actions improving the overall execution time. More critically, it is robust enough that it allows one to plan the sequence of actions offline and then execute them reliably on a Baxter robot.

\vspace{2mm}
\noindent Introduction video: 
\href{https://youtu.be/00kEztqytRU}{\texttt{\textcolor{blue}{youtu.be/00kEztqytRU
}}}.\\ Code: 
\href{https://github.com/DanManN/planning_baxter/tree/aggregation}{\texttt{\textcolor{blue}{github.com/DanManN/planning\_baxter/}}}

\end{abstract}
\section{Introduction}

Retrieving a target object from a messy and constrained space, such as taking out a bottle of water from a fridge, requires a robotic arm to relocate other objects blocking access. Humans routinely perform such tasks with a high degree of success, and the manipulation primitives used to execute such tasks are not limited by pick-and-place-based rearrangements. Instead, they often involve non-prehensile manipulation, such as pushing and pulling actions. Endowing robots with such skills is highly desirable, especially if they are tasked to carry out ordinary human tasks. Humans execute such manipulation operations, including grouping objects before performing a push, in a naturally robust manner, while keeping the number of actions low. 

\begin{wrapfigure}{r}{0.5\textwidth}
\vspace{-.3in} 
\centering
\begin{overpic}[width=0.49\textwidth]{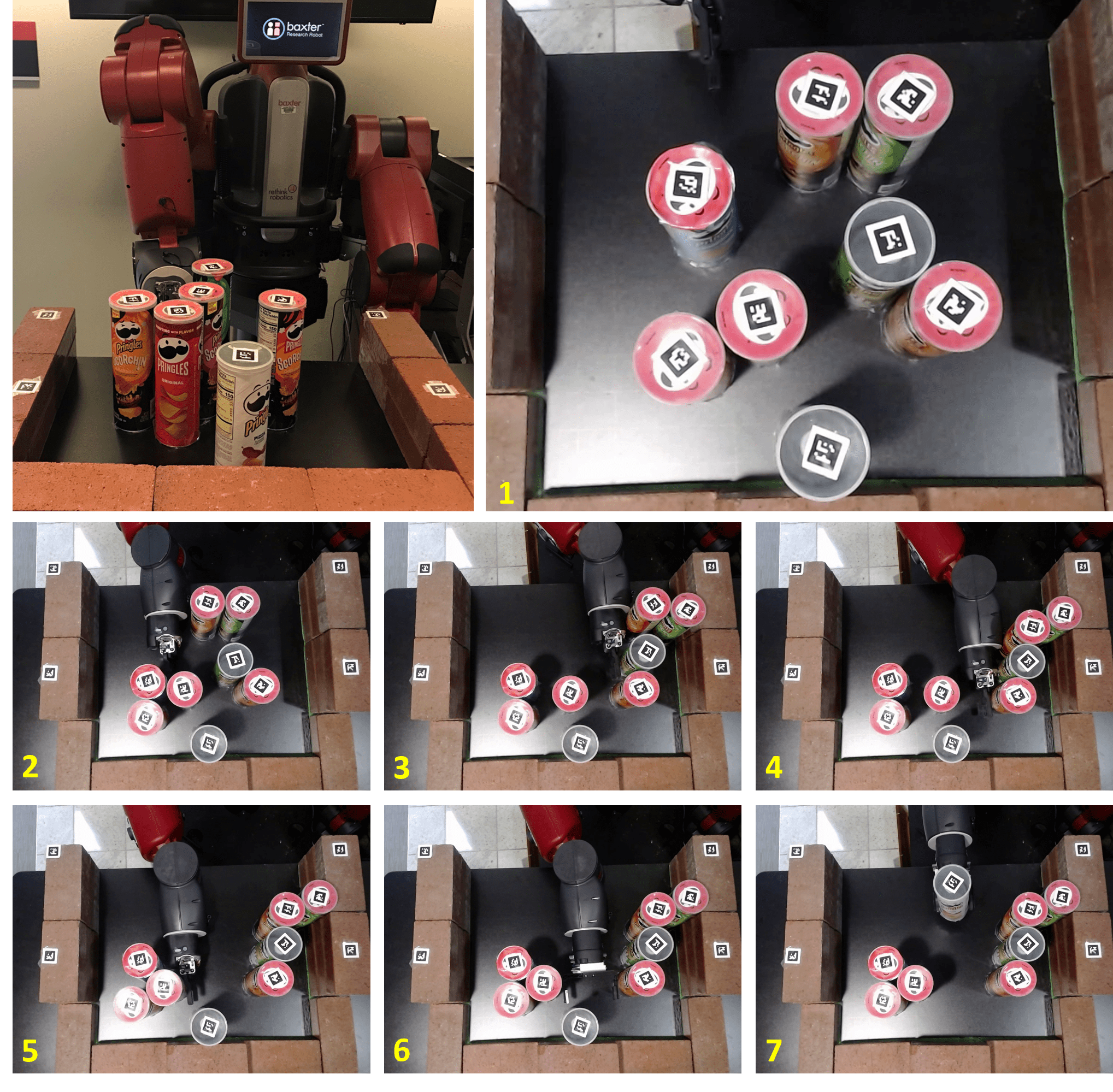}
\put(79.5,56){\textcolor{yellow}{\target}}
\end{overpic}
   \vspace{-.15in}
    \caption{\small Top left: the experimental setup. Top right: Scene S2 is one of the initial configurations for the experiments. The robot must move cluttered objects blocking access to a target object $\target$ placed at the bottom close to the wall. (2) to (5) demonstrate the pushing actions executed after a given plan by the proposed method. A simple grasping plan is performed from (6) to (7).}
    \label{fig:setup}
    \vspace{-.3in}
\end{wrapfigure}
\begin{wrapfigure}{l}{0.5\textwidth}
\vspace{-.3in}
    \centering 
    \includegraphics[width=0.5\textwidth]{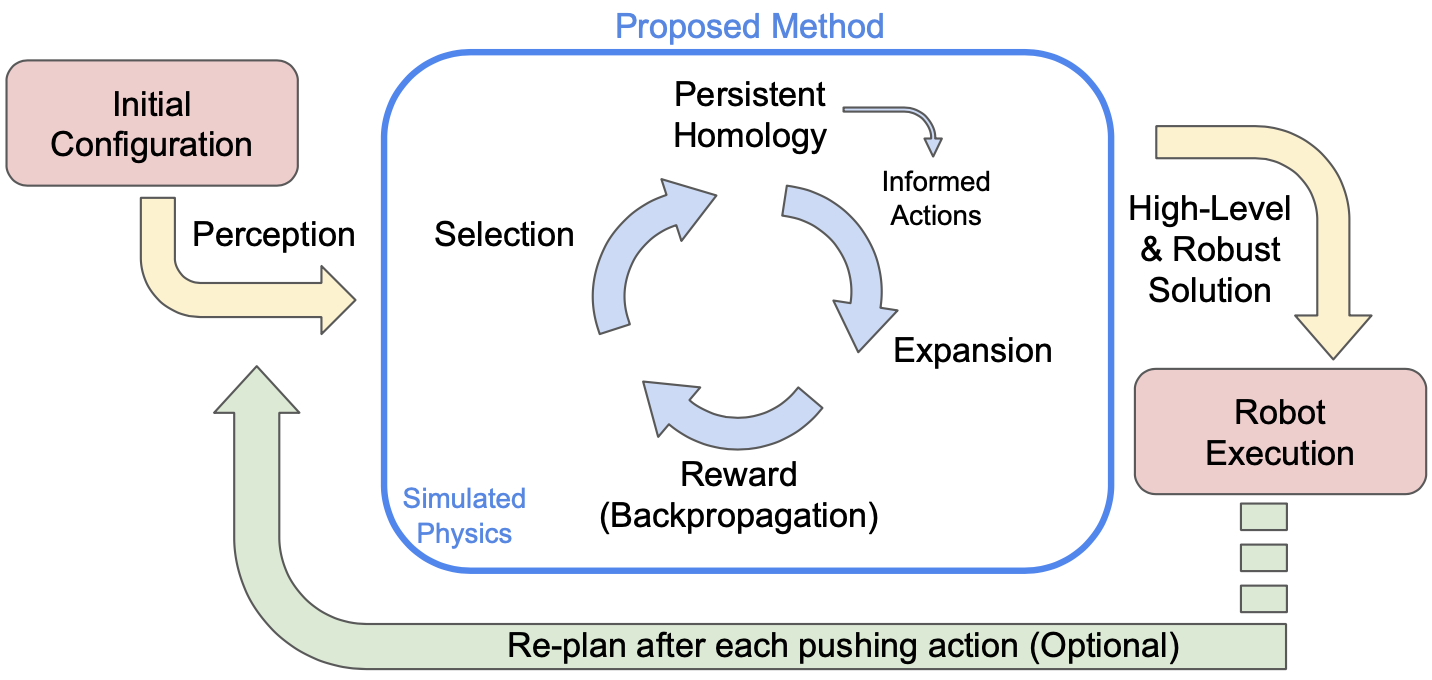}
    \vspace{-.35in}
    \caption{\small Proposed framework integrating Persistent Homology for informed actions/rewards and Monte-Carlo tree search.}
\vspace{-0.35in}
\label{fig:method}
\end{wrapfigure}

\noindent This work proposes strategies for solving object retrieval problems focusing on \emph{efficiency and robustness}. An example run is shown in Fig. \ref{fig:method}, in which the robot must retrieve the target object located at the lower portion of the workspace (a Pringles can of a specific flavor). Our study tackles the challenge by exploring two related sub-strategies: (i) clustering objects to inform the choice of pushing actions, and (ii) searching through and evaluating feasible pushing operations before executing them in the real world. The first is motivated by human behaviors of implicitly grouping objects before pushing them simultaneously. Topological tools, such as persistent homology, have been successfully employed to comprehensively identify manageable object clusters for selecting the proper pushing actions \cite{vieira2022ph}. 
The second direction leverages Monte-Carlo Tree Search (MCTS) to explore the feasible identified pushing actions with a high reward level in terms of the number of obstacles removed from the path to the target and how dispersed the clusters are after the performed actions. The reward metric is intuitive since a higher number of obstacles removed results in a higher chance to solve the task faster, and dispersing the clusters leads to easier pushing actions for consecutive steps.

The proposed framework integrates MCTS and the informed actions/rewards provided by Persistent Homology. It is referred to here as {\bf PHIM} ({\bf P}ersistent {\bf H}omology {\bf I}nformed actions and rewards for {\bf M}onte-Carlo tree search). For real-world experiments, the pipeline obtains the initial configuration from perception and then planning via PHIM to produce a high-level sequence of operations that result in successful execution reliably. The solution is executed on the Baxter robot. This can be done either by performing the plan open-loop or optionally re-planning after each pushing action execution.  Given the experimental evaluation, PHIM achieves a significantly higher success rate in solving cluttered problems in constrained, shelf-like workspaces than alternatives, due to its long-horizon planning capability and solution robustness. Furthermore, plans produced by PHIM demonstrate are visually more natural and human-like. 

In summary, this work contributes: 
    {\bf 1)} {\bf PHIM} ({\bf P}ersis\-tent {\bf H}omology {\bf I}nformed actions and rewards for {\bf M}onte-Carlo tree search) is a principled framework that integrates topological tools and Monte Carlo methods to achieve long-horizon planning for object retrieval via non-prehensile actions.

\noindent {\bf 2)} In comparison to existing methods, including methods that only employ topological tools but do not perform planning over multiple horizons, PHIM delivers significantly improved solution optimality in terms of the success rate and the number of actions taken. At the same time, PHIM incurs little extra computations cost.\\
\noindent {\bf 3)} PHIM demonstrated superior robustness to uncertainty (e.g., in comparison to \cite{vieira2022ph}, which lacks real robot experiments) as its plans can be directly executed on a real Baxtor robot with inherently motor uncertainty plus additional perception uncertainty. 

\label{sec:intro}

\vspace{-.2in}
\section{Related Work}
\vspace{-.15in}

Working in a constrained and cluttered environment requires manipulation tasks, such as pick-n-place and object rearrangement. Prehensile manipulation generally assumes accurate pose estimation of objects, which is challenging for cluttered setups with action uncertainty. In contrast, non-prehensile actions can be used in a cluttered and constrained workspace with uncertainty to rearrange multiple objects using a single operation, thus allowing large-scale object manipulation \cite{huang2019large}. Sometimes they are preferred over picking \cite{nieuwenhuisen2013mobile, shome2019towards, song2019object, pan2020decision, havur2014geometric} since pushing can be executed with simpler and smaller end-effectors for constrained workspaces. Non-prehensile actions, however, are less predictable and increase uncertainty regarding object placement; this motivates solutions for estimating the outcome of pushing actions \cite{zhou2019pushing, zhou2018convex, huang2021dipn}.

Previous works in reaching through clutter focused on recognizing objects to be relocated to allow a collision-free region to reach a target \cite{lee2019efficient, nam2019planning}. Uncertainty arising from occlusion or sensor noise complicates this problem, and efforts aim to minimize their effects by inferring object shape \cite{price2019inferring, wong2013manipulation}, reason about pose uncertainty \cite{wang2020safe}, or using probabilistic filtering
\cite{poon2019probabilistic}. To achieve higher success rates, human interaction has been proposed to guide a high-level plan, which proposes an ordered sequence of objects to be pushed to approximate intermediate goal positions \cite{papallas2020non}. For faster kinodynamic planning, non-prehensile actions are allowed to have quasi-static interactions \cite{haustein2015kinodynamic}. 

The proposed method takes advantage of persistent homology to inform the selection of efficient and robust pushing actions, which outperform, in terms of planning time and the number of actions, prior efforts in the same domain \cite{papallas2020non, haustein2015kinodynamic, vieira2022ph}. Moreover, it provides high-level solutions that are robust under uncertainty (pose, arm movements, non-prehensile manipulation of objects). Related problems, where the target has to be pushed to a desired goal, have been explored \cite{cosgun2011push}. Some approaches learn an optimal policy from visual input \cite{yuan2018rearrangement} or aim to compute effective manipulation state/action sequences \cite{haustein2019learning}.

\label{sec:related}

\vspace{-.2in}
\section{Problem Setup and Notation}
\label{section: problem setup}
\vspace{-.15in}

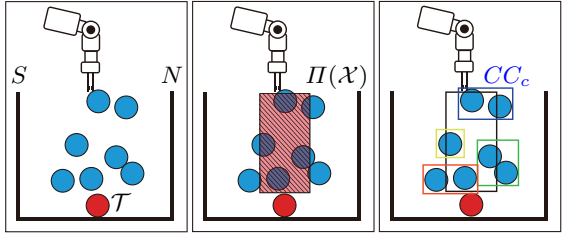
\begin{wrapfigure}{r}{0.6\textwidth}
\vspace{-.45in}
\begin{center}
\begin{overpic}[width=.6\columnwidth]{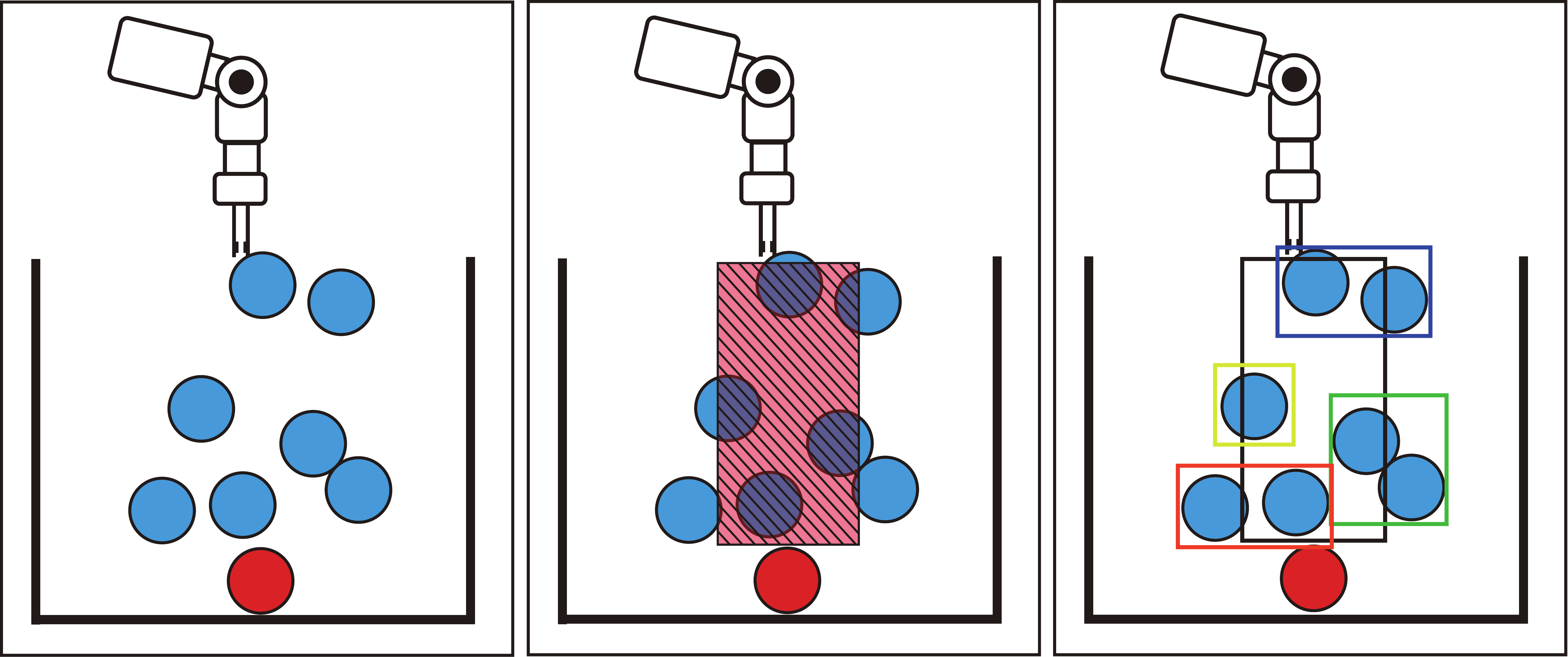}
\put(19,4){\small \target}
\put(28, 27){\small $N$}
\put(1, 27){\small $S$}
\put(86, 27){\small \textcolor{blue}{$CC_c$}}
\put(54, 27){\small $\pathregion(\config)$}
\end{overpic}
\end{center}
\vspace{-.3in}
\caption{\label{fig:pathregion}(left) The workspace $\workspace$ and the robot arm for the configuration of Fig. \ref{fig:setup}, where $\target$ is the target. (middle) The path region $\pathregion(\config)$. (right) Connected components in  $\pathregion(\config)$ using $r = 0.116$, where $CC_c = CC_c(\config, r)$ is the closest component to the gripper.}
\vspace{-.3in}
\end{wrapfigure}

Let $\workspace \subset \mathbb{R}^3$ be a bounded region with a cross-section given by a polygon and $\partial \workspace$ is its boundary, where {\it movable  obstacles} $\objects=\{o_1, \ldots, o_n\}$ and a target object $\target$ are approximated by cylinders residing inside $\workspace$. Assume that an arm with a gripper $g$ can reach a target object at any position in $\workspace$. Furthermore, the  arm can access the interior of $\workspace$ from only one face of $\partial\workspace$, where due to limited accessibility, it cannot perform overhand grasps or lift the objects. Collisions between $\partial \workspace$ and the arm are not permitted. In other words, the robot respects the workspace constraints and may move along a 2D plane below the center of mass of the objects to avoid toppling them.  
The grasping of the target $\target$ requires the arm's last link to be collision-free with all obstacles, including the boundary $\partial \workspace$. The robot is a Baxter with a 7-DoF articulated robotic arm with a gripper attached at its end. 
The task  is to reach and grasp the target object $\target$ with a minimal set of actions used to move obstacles blocking access to $\target$.

The configuration $\config$ is defined as $\config = \{ \positions, p_{\target}, s\} = \{ (p_1, \ldots, p_n), p_{\target}, s \} \in \workspace^{n+1}\times SE(2)$, where $p_i \in \workspace$ defines the object $o_i$'s  coordinates and $p_{\target}$  defines the $\target$'s coordinate, respectively, while $s$ is the position and orientation of the robot's gripper $g$. To simplify the notation, $\config[o_i]=p_i$ and $\config[\target]=p_\target$ represent that 
$o_i$ and $\target$ are at position $p_i$ and $p_\target$, respectively. $\config[g]=s$ represents that the gripper $g$ of the robot is at pose $s \in SE(2)$. Let $V(p)$ be the subset of $\workspace$ occupied by an object at position $p$. A configuration $\config\in \workspace^{n+1}\times SE(2)$ is \emph{feasible} if: no objects are overlapping; objects are not dropped, toppled, or deformed. The arm is allowed to push multiple objects simultaneously, and the resulting action must lead to a feasible configuration.

\vspace{-.2in}
\section{Method}
\label{sec:method}
\vspace{-.15in}

Persistent homology is a data analysis tool that provides topological features of a space at different spatial resolutions. It is often applied for analysis of point-cloud data (e.g., see \cite{MR3328629}, and \cite{KaczynskiTomasz2004Ch/T}). Briefly, it considers expanding balls of radius $r$ centered at each point of a collection of points. As the balls grow, track the unions of all these balls as they overlap each other, given a $1$-parameter family of spaces. For each radius $r$, one can build a Vietoris–Rips complex (abstract simplicial complex) using the information given by the intersection of the $r$-balls (for more details, see \cite{KaczynskiTomasz2004Ch/T}). In this setting, persistent homology is the homology of the Vietoris–Rips complex as a function of $r$. Intuitively, persistent homology counts the number of connected components \red{(object clusters)} and holes of various dimensions and keeps track of how they change with parameters.

\begin{wrapfigure}{l}{0.23\textwidth}
     \centering
    \vspace{-.2in}
    \begin{overpic}[width=0.23\textwidth]{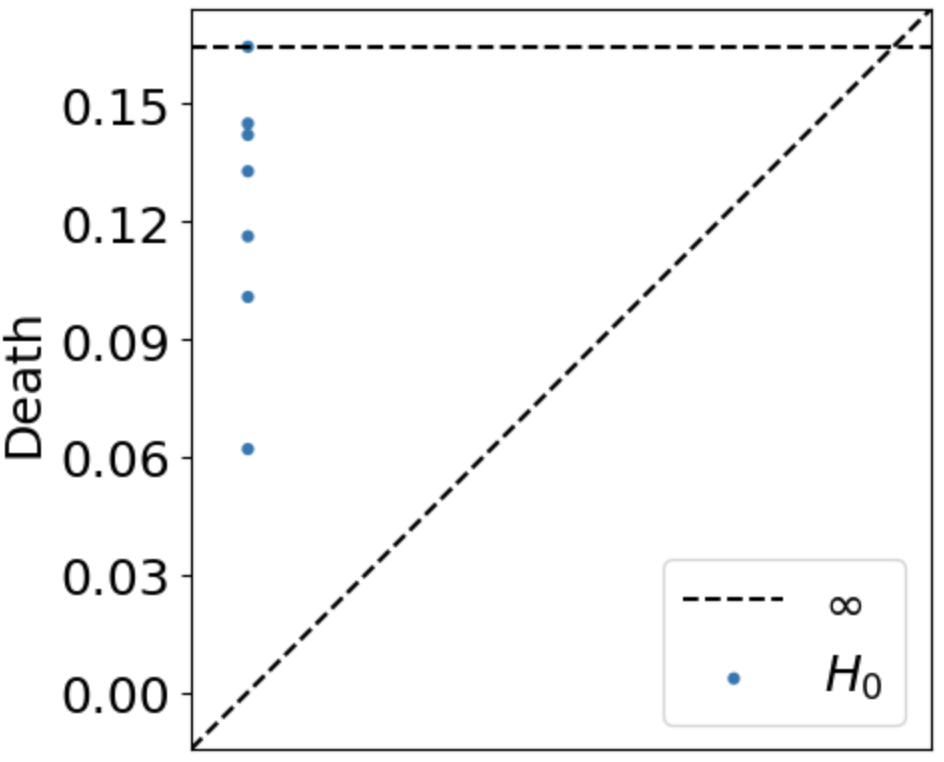}
    \end{overpic}
   \vspace{-.3in}
    \caption{\small The persistence diagram for the positions of the obstacles of Fig. \ref{fig:setup}.}
    \label{fig:pers_diagram}
    \vspace{-.3in}
\end{wrapfigure}

\begin{wrapfigure}{r}{0.5\textwidth}
    \centering
    \vspace{-.3in}
\begin{overpic}
[width=0.5\columnwidth,tics=5]{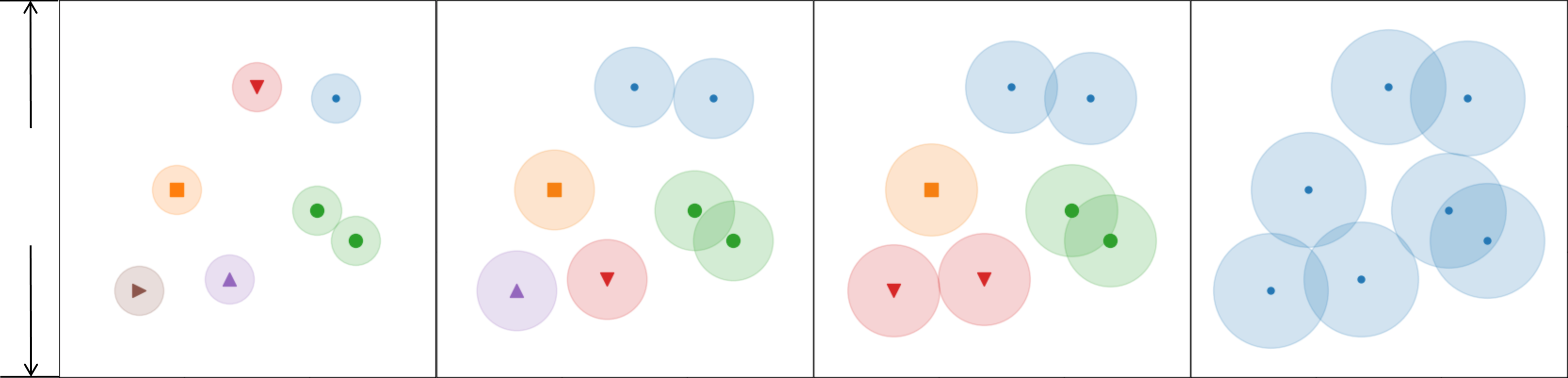}
\put(0.8, 10){\rotatebox{90}{{\small $0.5$}}}
\end{overpic}
    \vspace{-.28in}
\caption{\small Examples of clusters (connected components) that persist obtained for four different radii $r$ in the persistence diagram in Fig. \ref{fig:pers_diagram}. (left) For $r = 0.062$, there are 6 different clusters shown by different markers.  (left-middle) For $r = 0.1$, there are 4 clusters. (right-middle) For $r = 0.116$, two clusters. (right) Only one cluster for $r \geq 0.144$.}
    \label{fig:connec_comp}
        \vspace{-.3in}
\end{wrapfigure}

\noindent To find clustering that persists, it is sufficient to focus on the zeroth-homology (zero-dim. homology) where the coefficient $\mathbb{Z}_2$ is ideal for computing connected components (clusters). While the radius $r$ increases, the zero-dimensional persistent homology stores when the ball in one connected component first intersects a ball of a different connected component, merging both into one, see Fig. \ref{fig:connec_comp}.

\noindent {\bf Non-Prehensile Manipulation Approach} Given a target $\target\in \objects$, denote $\pathregion(\config)\subset \workspace$ as the {\it path region}, where the robot arm will push the obstacles to clear the path to the target. The idea is to use persistent homology to find clusters (connected components) inside $\pathregion(\config)$ and use this information to push objects. Since $\pathregion(\config)$ is smaller than the workspace, the arm has to fit inside $\pathregion(\config)$ to perform the action. Thus the shape and size of $\pathregion(\config)$ depends on $\workspace$ and the width $w$ of the arm. For simplicity, $\workspace$ is considered to be a rectangular shelf described by the parallel walls of the shelf $N$ and $S$ as in Fig \ref{fig:pathregion}.

To describe $\pathregion(\config)$, first select axis $x$ to be associated with the depth of the shelf and $y$ to correspond to the width of the shelf with $y=0$ and $y=w$ for points at the south wall $S$ and north wall $N$, respectively, where $w$ is denoted by the width of the shelf. Then, $\pathregion(\config)$ is defined to be the rectangle given by $(\config[g]_x, \config[\target]_y - w)$ and $(\config[\target]_x, \config[\target]_y + w)$, where $\config[\target] = (\config[\target]_x, \config[\target]_y)$ and $\config[g] = (\config[g]_x, \config[g]_y)$, see Fig. \ref{fig:pathregion}. When the target is too close to the walls $S$ and $N$, it is possible to apply a planar rotation matrix $R_{-\phi}$ to obtain $\pathregion(\config)$ with incidence angle $\phi$ as in \cite{vieira2022ph}.

Having $\pathregion(\config)$, obstacles can be clustered by applying persistent homology inside $\pathregion(\config)$. Define
$\connectedcomp(\config, r)$ to be the collection of connected components (clusters) for the radius $r>0$. It is enough to use the zero-dimensional persistent homology to obtain the connected components and their generators. The goal is for the arm to try to push all obstacles in the closest connected component, denoted by $\connectedcomp(\config, r)_c$, to the griper $\config[g]$. The performed action $a$ is successful when all obstacles in $\connectedcomp(\config, r)_c$ are removed from $\pathregion(\config)$.

 Due to the unpredictability of a non-prehensile action $a$, it is advantageous to bestow a reward on each action performed based on the number of obstacles removed from the path region and the number of connected components created by the pushing action $a$. The reward function $\reward$ in MCTS for a given action $a$ is defined by $\reward(a) = b(a) + t(a)\cdot m(a),$
where $b(a) = \#\pathregion(ch(\config, a)) - \#\pathregion(\config)$ is the number of obstacles removed from the path region; $m(a) = \mathrm{max}(\#\connectedcomp(ch(\config, a), r) - \#\connectedcomp(\config, r), 0)$ is the number of connected components created by the action $a$; and
$t(a)=1$ if $b(a)>0$ or $0$ elsewhere. 
Note that the term $m(a)$ adds a positive reward for the configuration that increases the number of clusters (connected components), resulting in a configuration of sparse obstacles that leads to subsequent easier pushing actions.

The pushing actions are based on the circumscribed rectangle that contains $\connectedcomp(\config, r)_c$, denoted by $\rect(\config, r)$. The actions are either by a sweeping movement from the bottom of $\rect(\config, r)$  to the top of $\rect(\config, r)$ or from top to bottom of $\rect(\config, r)$, denote the action by $\mathsf{up}$ and $\mathsf{down}$, respectively.

Observe that for a given configuration $\config$, each radius $r>0$ may define a different $\rect(\config, r)$, and without a proper discretization of the radii, the tree will grow exponentially. The persistence diagram provides an informed manner to select radii that lead to robust actions and bounds the exponential growth. The radii that present such properties are the collection of radii that persist under a selected hyper-parameter $\nu$, more specifically, a {\it persistent radius} for a given parameter $\nu>0$ is a radius $r>0$ such that the number of connected components in the persistence diagram between $r$ and $r+\nu$ remains the same. Denote $R_\nu$ by the set of all persistent radii for a given value $\nu>0$. In the persistence diagram in Fig. \ref{fig:pers_diagram}, $R_{0.015}=\{0.062, 0.1, 0.116, 0.144\}$. Note that $R_\nu$ always contains the radius where the last connected component dies, thus $R_\nu\neq \emptyset$. 

Another important parameter is the gripper width $h$, where for radii less than $h$ the gripper cannot position between two clusters, thus $R_\nu$ can be further decreased by removing all radii less than $h$ from it, denote this new collection of radii by $R(\config)=R_{\nu,h}(\config)$. For example, in the persistence diagram of Fig. \ref{fig:pers_diagram}, $R_{\nu, h}(\config)=\{0.062, 0.1, 0.116, 0.144\}$ for $\nu = 0.015$ and $h = 0.05$. In \cite{vieira2022ph}, there are two proposed algorithms: PHIA takes the minimal radius of $R_{\nu, h}$ and performs the pushing actions, and PHIS builds a tree with nodes being the configurations and propagates by performing the pushing actions for each radius in $R_{\nu, h}$ then it selects the path that leads to success and minimizes the number of actions. 

Given a configuration $\config$, the set of all possible actions is represented by $R(\config)\times \{\mathsf{up}, \mathsf{down}\}$, hence the expansion stage of MCTS for $\config$ produces children $ch(\config, a)$ where $a=(r,d)\in R(\config)\times \{\mathsf{up}, \mathsf{down}\}$, $r$ a radius and $d$ a direction. Actions $a$ may result in failure either by dropping/toppling obstacles or pressing obstacles against the wall. In all cases, a child node obtained by unfeasible action is marked as a failure, and no reward is given. The framework that integrates the informed actions and rewards by Persistent Homology and MTCS is named in short by {\bf PHIM} ({\bf P}ersistent {\bf H}omology {\bf I}nformed actions and rewards to {\bf M}onte-Carlo tree search). For sample pushing actions planned by PHIM, see Fig. \ref{fig:setup}.


\vspace{-.2in}
\section{Experiments}
\label{sec:experiments}
\vspace{-.12in}

\begin{wrapfigure}{r}{0.5\textwidth}
\vspace{-0.5in}
    \centering
    \begin{overpic}[width=0.49\columnwidth]{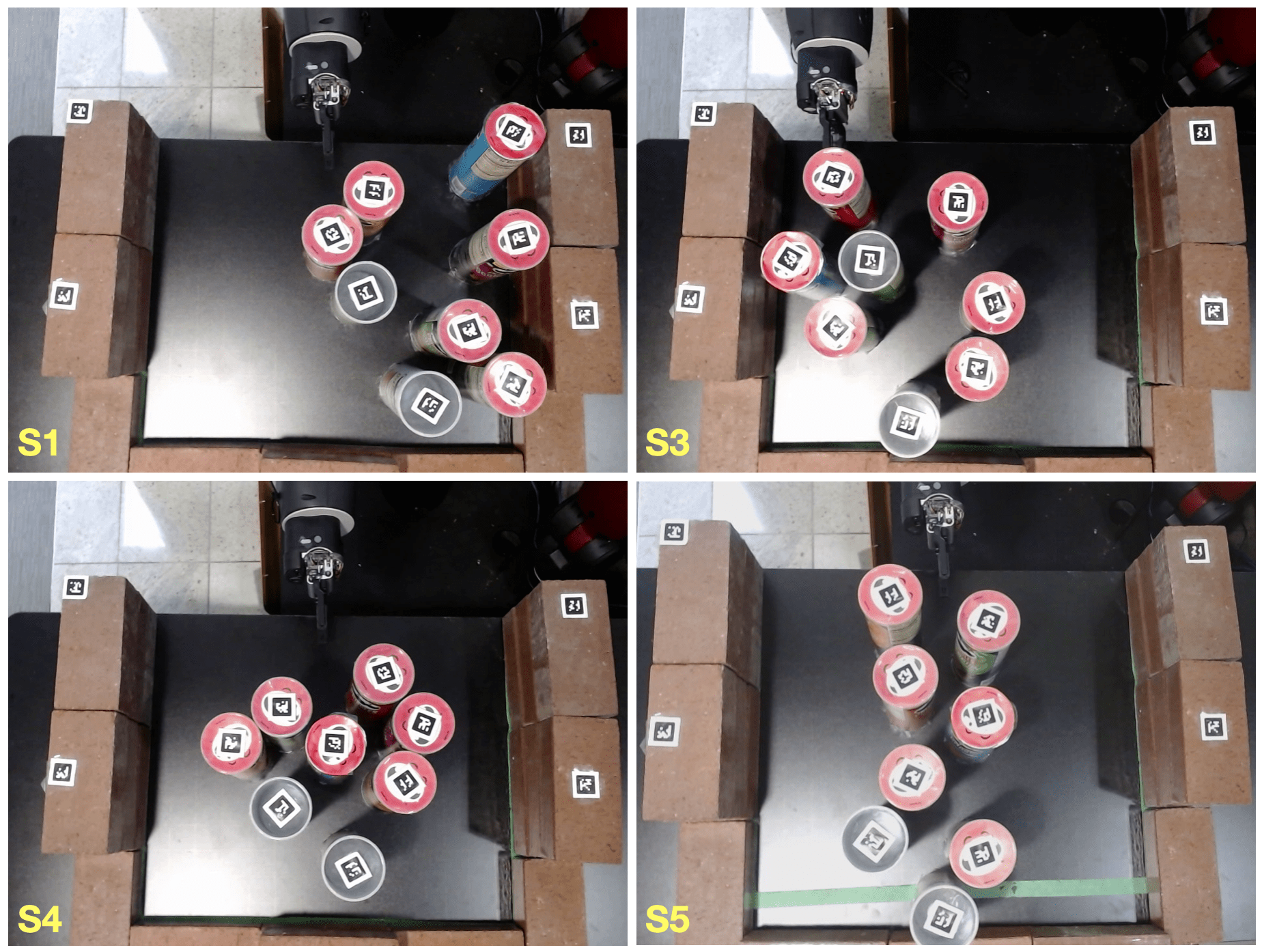}
    \end{overpic}
    \vspace{-0.1in}
    \caption{\small (top row) the randomized baselines fail to find a plan; (bottom left - S4) GRTC-H fails to find a plan, and OOA fails to execute the plan;  (bottom right - S5) GRTC-H, OOA, and PHIA perform the most actions. Scene S2 is given in Fig.~\ref{fig:setup}.}
    \label{fig:scenes}
    \vspace{-0.25in}
\end{wrapfigure}

\begin{wrapfigure}{l}{0.5\textwidth}
\vspace{-0.3in}
    \centering
    \begin{overpic}[width=0.24\columnwidth]{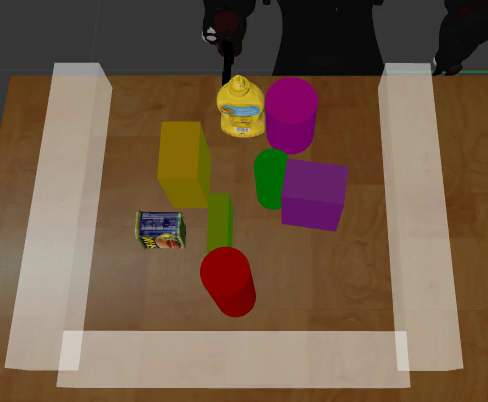}
    \put(0,1){\textcolor{yellow}{M2}}
    \end{overpic}
    \begin{overpic}[width=0.24\columnwidth]{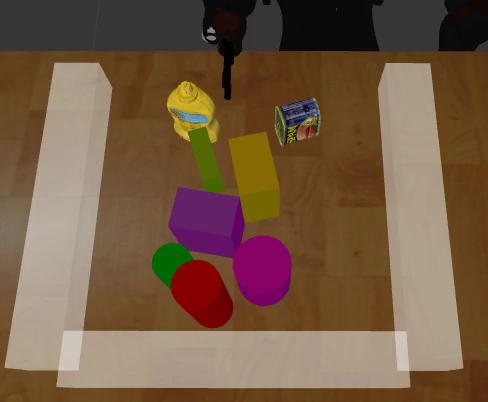}
    \put(0,1){\textcolor{yellow}{M5}}
    \end{overpic}
    \vspace{-0.15in}
    \caption{\small Randomized baselines M2 and M5 in simulation  with varying shape obstacles.}
    \label{fig:simulated_scenes}
    \vspace{-0.3in}
\end{wrapfigure}

To evaluate the proposed method, 25 real-world and 50 simulated experiments were performed and compared against baselines as described below. The real-world scenarios are described by S1, S2, S3, S4, and S5 in Fig. \ref{fig:scenes} and \ref{fig:setup}. The simulated experiments are labeled M1 through M10 as depicted in Fig. \ref{fig:simulated_scenes}.

\noindent {\bf Experimental setup:} The simulated experiments were run on a Ubuntu workstation with {\it Intel Core i5-8259U 3.8Ghz 16GB RAM}. Gazebo was used for simulating the actions, and MoveIt was used for forming motion planning queries to the Bi-EST planner \cite{hsu1997path} in OMPL.

\noindent For real-world experiments, the Baxter robot was used to perform the pushing actions and grasp the target. The choice of Baxter was ideal for the real experiment since it performs the moving actions with a small degree of imprecision, which helps to show the robustness of the proposed approach. An overview of the workspace is provided by an Intel Realsense D435 RGB-D camera. For each object, its position is detected with a 2D fiducial marker at the top using Chilitags.

\noindent {\bf Choice of parameter:} 1 target and 7 obstacles with enough height to have the center of mass above the contact points performed by the pushing actions of Baxter's arm. Bricks are used to create artificial walls to simulate a shelf workspace with a desirable size, and it allows recording and running perception over the table. The gripper can lift the target a bit to perform the retrieve action. However, it is not allowed to retrieve while passing over the obstacles. The objects have enough weight to provide friction and simulate the real grasping.

The hyper-parameter $\nu$ for the persistent homology algorithms is selected to be $0.015$m, the margin of error for the arm's pushing actions, which is empirically chosen to be the difference between the desirable position minus the actual position after performing the actions.

Another useful hyper-parameter to consider is the width of the gripper $h$. The selection of $h$ dictates if the gripper can be successfully fitted between two clusters (connected components). During the experiments, $h$ is considered the real width plus an extra margin of error. Thus, $h$ is selected to be greater than $0.03$ + $0.015$m ($h=0.05$m). Furthermore, for the exploration parameter in the Monte-Carlo tree search, the conventional choice of $\sqrt{2}$ is chosen. The overall planning time threshold for all methods is 500 seconds. Above this threshold, failure is reported and real-world experiment is not performed.

\textbf{Collisions:} Soft collisions between obstacles and walls are allowed. Nevertheless, the simulated actions prevent hard collisions and the arm pressing the obstacles against the wall. In fact, actions that resulted in the arm pressing the obstacles were the only issue presented when the simulated experiments were performed in real-world scenarios.

\begin{wrapfigure}{r}{0.6\textwidth}
\vspace{-0.25in}
\centering
\begin{overpic}[width=0.59\columnwidth]{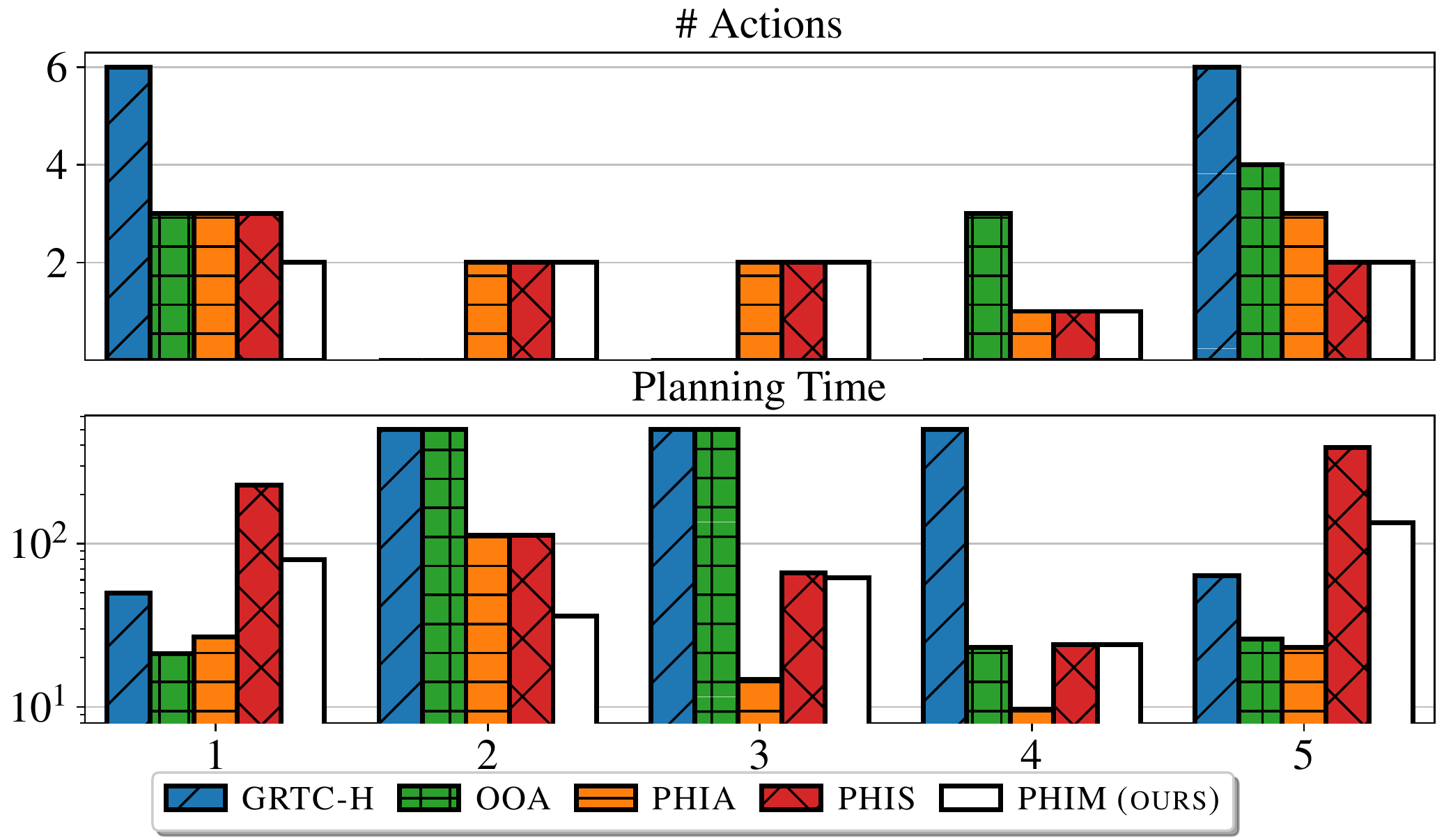}
\end{overpic}
\vspace{-0.1in}
\caption{\label{fig:exp_results} 
\small Real-world actions (top) and planning time (bottom) performed by each method for scenes S1 to S5 of Fig. \ref{fig:setup} and \ref{fig:scenes}, for GRTC-H, OOA, PHIA, PHIS, and PHIM (ours), where applicable. All PHI* methods do reasonably well; PHIM always provides the best solution quality.}
\vspace{-0.3in}
\end{wrapfigure}
It was surprising that the challenge of avoiding dropping and toppling was easily solved. The proposed solution to fix pressing objects against the wall is to consider the walls in a simulated environment as movable objects with five times the weight of one obstacle. Any action resulting in moving the wall is considered unfeasible. Incorporating this extra condition increased the chances of failing during the planning phase. However, experiments showed that the PHIM approach could find high-level solutions with minimal pushing actions without significantly increasing planning time.

{\bf Comparisons:} The proposed PHIM is compared against PHIA, PHIS, OOA \cite{vieira2022ph} and the modified GRTC-Heuristic \cite{papallas2020non}. All algorithms were adapted to prevent the arm from pressing the objects against the wall. Without this extra condition, PHIA, OOA, and GRTC-Heuristic fail in all real-world experiments. This additional check condition does not require changing the algorithms. The algorithm OOA is a simple approach to push each obstacle one by one in the path region $\pathregion(\config_0)$ of the initial configuration $\config_0$. And modified GRTC-Heuristic (GRTC-H) is based on the algorithm used in \cite{papallas2020non} that pushes obstacles in a straight line to their goal region outside of the path region $\pathregion(\config_0)$, where the goal is randomly selected to be outside of $\pathregion(\config_0)$ without overlapping any other obstacles. Both algorithms are randomized approaches to execute pushing actions without using clustering information from the persistent homology.

{\bf Robustness:} To evaluate the robustness of the proposed method, the real-world experiments had the position of each object slightly different from the positions given to the planner to solve\textcolor{black}{, adding a noise of 3cm to the tag-based position}. Moreover, the planner simulated the whole collection of pushing actions to solve the retrieval problem without re-planning after every step. In other words, the planner solved the problem in a simulated world in an offline manner. This choice of offline planning is motivated by the robustness of the algorithms based on persistent homology.

{\bf Discussion:} The success rate for GRTC-H is comparatively low, mainly due to taking a long time planning feasible actions (see Fig. \ref{fig:exp_results}). The planning time for OOA is the second fastest as it does not perform a search and only finds the closest obstacle to the gripper. However, it has a high chance of failing during execution as it is not robust. When it pushes an obstacle and if another one is near, it can push them together, thereby changing the  configuration for the next action. Thus, it results in a lower success rate.

\begin{wrapfigure}{l}{0.6\textwidth}
\vspace{-.3in}
    \begin{center}
\begin{overpic}[width=0.59\columnwidth]{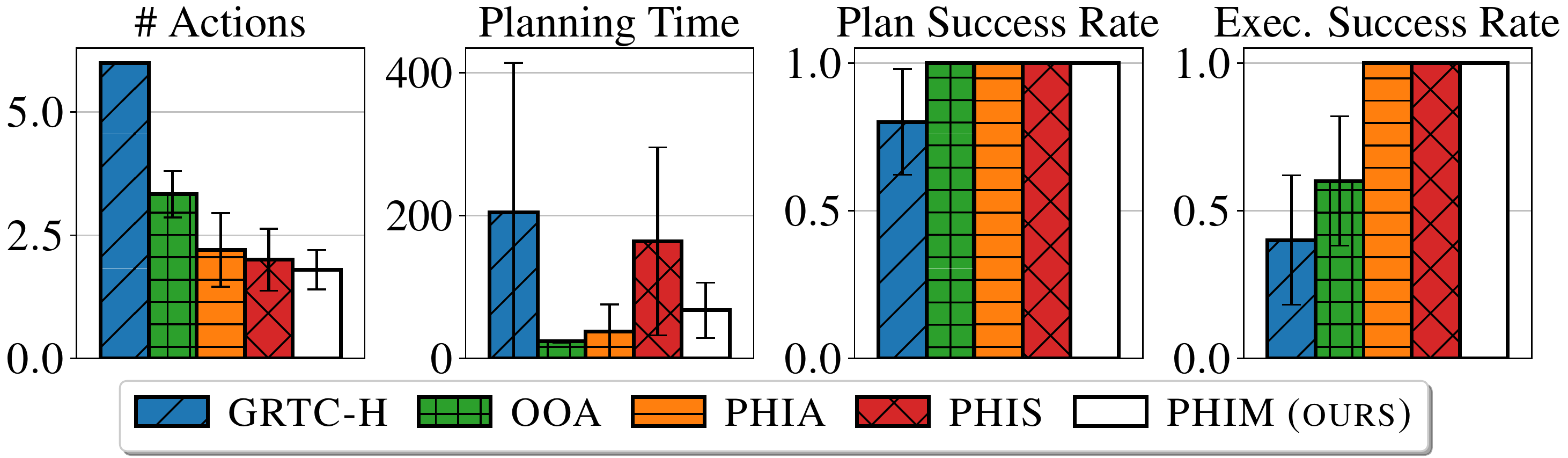}
\put(0,23){{\bf a)}}
\end{overpic}
.
\begin{overpic}[width=0.59\columnwidth]{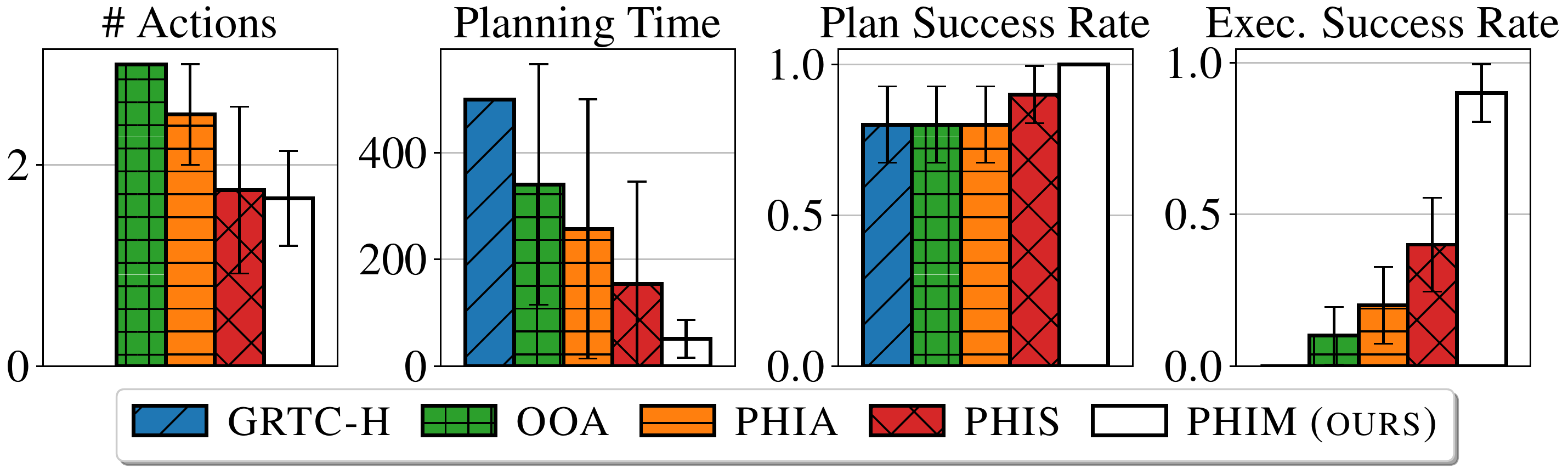}
\put(0,28){{\bf b)}}
\end{overpic}
\end{center}
\vspace{-.3in}
\caption{\small \label{fig:avg} 
Left to right: average number of actions, planning time, planning success rate, and execution success rate for {\bf a)} real-world and {\bf b)} simulated experiments. PHIM provides the best solution quality and much faster than PHIS. 
}
\vspace{-.3in}
\end{wrapfigure}

\noindent PHIM plans fewer actions than the other methods and has a $100\%$ success rate either in simulation or in real-world execution per Fig. \ref{fig:avg}. It achieves such performance without substantially increasing the planning time, showing that the rewards based on information collected by persistent homology are quite useful. In summary, when comparing all methods, PHIM accomplishes the most robust actions as well as reduces the number of actions needs to solve the problem without significantly increasing the planning time.

\vspace{-.2in}
\section{Conclusion}
\vspace{-.15in}

This paper introduces a robust and efficient method for planning non-prehensile manipulation actions in clutter. It uses topological tools and MCTS to obtain high-quality solutions for object retrieval problems. The solutions are robust under perturbations and gracefully handle the uncertainty associated with objects' poses and arm actions. Furthermore, it allows to successfully plan offline actions using physics simulations that are successfully executed online.  Real-world experiments show that PHIM has a higher success rate and is faster in finding robust solutions than alternatives. It reduces the number of pushing actions required and decreases planning time by avoiding the consideration of actions that result in time-expensive simulated tasks.

The proposed framework can be directly applied to highly dense clusters of objects. However, such a scenario is expected to perform similarly to the other methods since highly dense setups have a trivial topological shape to explore (fewer ways to cluster). A possible direction to be explored is to select obstacles to be removed by pick-n-place tasks, resulting in easier follow-up pushing actions to clean the path to the target.

\vspace{-.25in}

%
%
%
\bibliographystyle{splncs04}
\bibliography{bib/c}
%




\end{document}